\documentclass[conference]{IEEEtran}

\usepackage{times}
\usepackage{latexsym}
\usepackage{graphicx}
\usepackage{adjustbox}
\usepackage{float}
\usepackage{hyperref}
\usepackage{soul,color}
\usepackage[justification=centering]{caption}

\usepackage{fancyhdr}
\usepackage{lastpage}

\begin{document}


\title{What factors influence the popularity of user-generated text in the creative domain? A case study of book reviews}

\author{\IEEEauthorblockN{ Salim Sazzed} \\
\IEEEauthorblockA{\textit{Department of Computer Science,}
\textit{Old Dominion University,}
Norfolk, VA USA}

\IEEEauthorblockA{\textit{Department of Computer Science,}
\textit{University of Memphis,}
Memphis, TN, USA \\ saim.sazzed@gmail.com}
} 



\maketitle
\begin{abstract}
This study investigates a range of psychological, lexical, semantic, and readability features of book reviews to elucidate the factors underlying their perceived popularity. To this end, we conduct statistical analyses of various features, including the types and frequency of opinion and emotion-conveying terms, connectives, character mentions, word uniqueness, commonness, and sentence structure, among others. Additionally, we utilize two readability tests to explore whether reading ease is positively associated with review popularity. Finally, we employ traditional machine learning classifiers and transformer-based fine-tuned language models with n-gram features to automatically determine review popularity. Our findings indicate that, with the exception of a few features (e.g., review length, emotions, and word uniqueness), most attributes do not exhibit significant differences between popular and non-popular review groups. Furthermore, the poor performance of machine learning classifiers using the word n-gram feature highlights the challenges associated with determining popularity in creative domains. Overall, our study provides insights into the factors underlying review popularity and highlights the need for further research in this area, particularly in the creative realm.

\end{abstract}

\section{Introduction}
User-generated content has an indispensable role in circulating information regarding products and various other types of entities on social networks and e-commerce sites \cite{o2010using,hong2017understanding}. As predicting review usefulness has practical implications,  in this study, we scrutinize how various kinds of features of a review are connected to its usefulness. However, instead of exploring review usefulness, or, more precisely helpfulness for the e-commerce domain (e.g., product review) that has been studied by many earlier studies \cite{hong2017understanding,lutz2022longer,huang2015study,malik2018analysis}, here, we analyze book reviews that represent the creative domain. Note that exist broad distinctions between helpful product reviews in the e-commerce domain and the popularity of the book review. Although some elements may overlap in both types, the core components of book reviews are likely to be different than those of e-commerce product reviews. In the former, a reviewer narrates and describes his feelings and opinions toward various aspects of the book, such as the story, plot, and characters influenced by the reviewer's writing style (i.e., elements of creativity). In contrast, a product review depicts the features and functionality of the product and may even provide a comparison with the other products, which is highly unlikely in book reviews. On the feedback level, in the e-commerce domain, the potential customer may label a review as helpful if the product pros or cons are available in the review, which is not the point for the creative domain; rather, the writing style may be concomitant to the popularity.






As exploring the factors underlying the perceived popularity of user-generated text in the creative domain can provide insights for content creators and platforms, contribute to knowledge in the creative realm, and identify challenges that warrant further research, here we explore a set of book reviews. By analyzing a set of 2259 popular and 2555 non-popular book reviews, we aim to uncover any association between the popularity of a review and its content features. Note that the popularity of a review specifies how a user review is accepted by fellow readers, whether they like it or not. Various factors,  including review helpfulness, clarity, readability, and information and sentiment conveyed, among others, may help determine whether an individual likes a particular review or not.
We examine various affective and linguistic features and perform readability tests. Various features, such as the presence of sentiment and emotion, usage of common and unique words, connectives, and mentions of characters, are studied. Besides, we analyze the review length and readability of the reviews using two well-known readability tests, FRE (Flesch Reading Ease) and FKR (Flesch–Kincaid readability). Finally, we employ several traditional machine learning classifiers and transformer-based fine-tuned language models for classifying reviews into popular and non-popular groups. We observe that although some of the features, such as emotion or review length, demonstrate some distinguishing capabilities, most of them do not exhibit any noticeable difference between the two groups in seclusion. The FRE and FKG readability tests, which are based on statistical syntactic attributes, do not reveal any discernible variations between the two groups. Classifiers show poor performances utilizing unigram and bigram-based word features for this binary-level popularity classification task (macro F1 scores between  0.60 - 0.65), which suggests determining the popularity of book reviews is a highly complex task.

\textbf{Contributions:}

The main contributions of this study can be summarized as follows-

\begin{itemize}
    \item  We investigate diverse types of textual attributes in book reviews to find their relationships with attained popularity.
   
     \item We employ two readability tests, FRE and FKG, to find whether reading ease is connected to the popularity of the book review. 
     
    \item We study the performance of various machine learning classifiers and transformer-based language models for the automatic classification of popular and non-popular book reviews.
 
\end{itemize}




\section{Related Work}
Various aspects of user-generated reviews, including helpfulness, sentiment, and linguistic traits, have been investigated in many earlier studies \cite{hong2017understanding,sazzed2021hybrid,sazzed2022stylometric,huang2015study,sazzed2021ssentia,sazzed2022understanding}. Similar to review helpfulness, review popularity is crucial for understanding the significance of reviews. Although the two terms, review helpfulness and review popularity, are closely related concepts, differences exist between them \cite{wu2017review}. The various attributes of reviews may have contrasting relationships with review popularity and helpfulness. As mentioned earlier, existing studies related to review content analysis mainly focused on predicting review helpfulness in the e-commerce domain \cite{hong2017understanding,lutz2022longer,huang2015study,malik2018analysis}. The findings of these works are often inconsistent, as reported by Hong et al. \cite{hong2017understanding}. Lutz et al. \cite{lutz2022longer} found that longer reviews (e.g., Amazon customer reviews) with frequent shifts between positive and negative arguments are perceived as less helpful. Malik and Hussain \cite{malik2018analysis} utilized six machine learning classifiers and three real-life Amazon review datasets in their study. They discovered that the number of syllables, spaces, auxiliary verbs, and driving words in review text and reviewer's productivity score are effective predictors of a review helpfulness. Huang et al. \cite{huang2015study} discovered a word count threshold beyond which the word count negatively affects the helpfulness of a review. Krishnamoorthy \cite{krishnamoorthy2015linguistic} used linguistic features from reviews, alongside metadata, subjectivity, and readability features, to predict helpfulness. Experiments on two real-world datasets showed the efficacy of proposed approach. Eslami et al. \cite{eslami2018online} noticed that most helpful online reviews are often associated with medium length, lower review scores, and negative or neutral argument frames. Moreover, their research revealed that review length is one of the most important factors in predicting the helpfulness of an online consumer review. Hong et al. \cite{hong2017understanding} performed a meta-analysis on the findings of the influence of determinants on review helpfulness. The authors found most of the review-related determinants (e.g., review depth, review readability, linear review rating, review age) and two reviewer-related determinants (i.e., reviewer information disclosure and reviewer expertise) have inconsistent conclusions( i.e., perceived review helpfulness) based on existing literature. They found several factors such as the criteria for helpfulness determination, the source of the reviews, and varied product domains are responsible for most of the mixed findings. Chua and Banerjee \cite{chua2015understanding} investigated the relationship between the review helpfulness with respect to reviewer reputation, review rating, and review depth. Utilizing Amazon review data, the authors found a positive correlation between the review helpfulness and the reviewer profile and review depth. However, when rating is concerned, they observed negative relations. Besides, they also noticed reviews having harsh criticism and adequate depth are usually received positively by the user.




\section{Dataset}
The review corpus used here consists of a collection of Goodreads book reviews obtained from the Kaggle website \footnote{\url{https://www.kaggle.com/datasets/notkrishna/top-100-science-fiction-books-and-their-reviews}}. Originally, it consists of around 10,000 reviews written towards top  100 science fiction books (ranked based on Goodreads ratings). However, upon examination, we find that a significant number of reviews comprise various types of issues, including missing values for the review text and like count, as well as duplicate sentences.
We remove duplicate sentences from the reviews and exclude problematic reviews with missing fields. After clean-up, the  corpus consists of 8845 reviews \footnote{\url{https://github.com/sazzadcsedu/ReviewPopularity}}.

\begin{figure*}
\includegraphics[width=.92\linewidth]{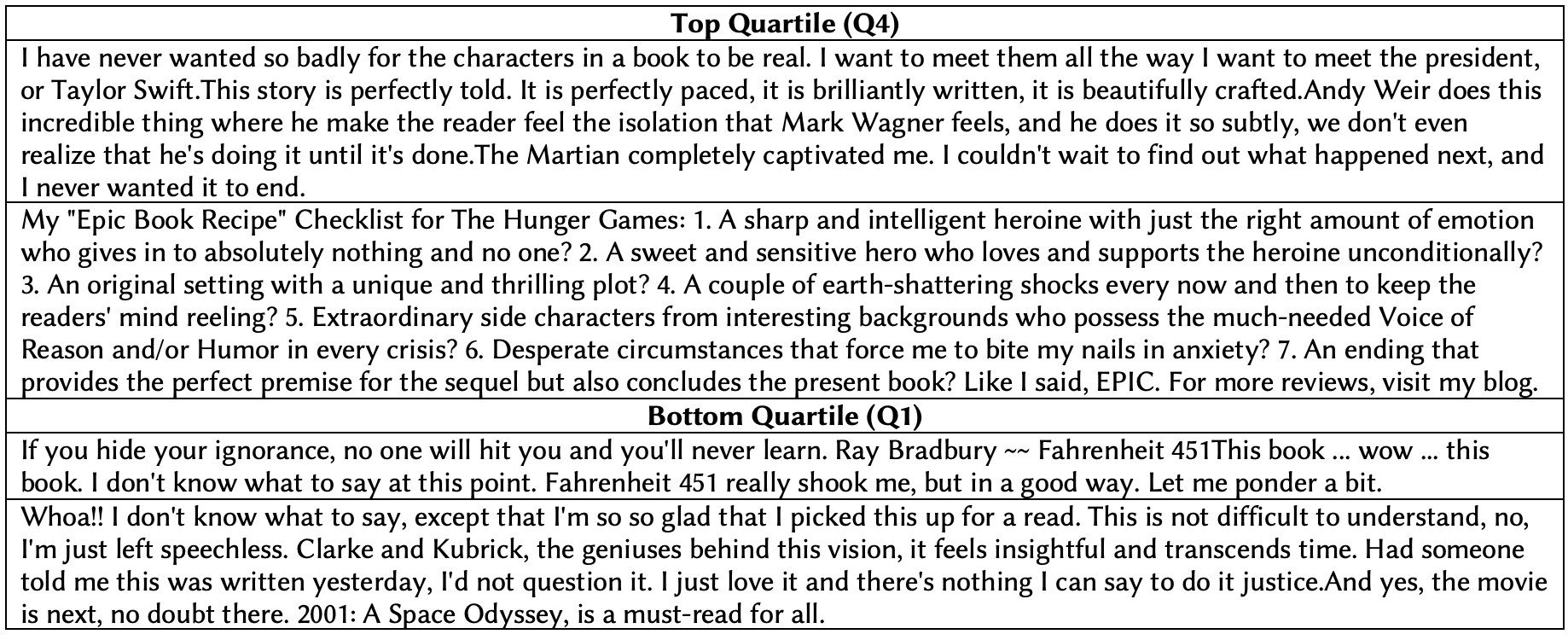}
\centering \caption{Examples of books reviews from popular (Q4) and non-popular groups (Q1)}
\label{examples}
\end{figure*}

As each book has different distributions of the \textit{like} count, it is more reasonable to decide the popularity of a review based on the \textit{like count} distribution of the individual book than considering all the books in the corpus. We split reviews of each book into four quartiles (Q1-Q4) based on the \textit{like count}.  The reviews of Q4 (i.e., like count above 75\% percentiles) and Q1 (i.e., like count below 25\% percentiles) are considered popular and non-popular reviews, respectively. Our initial analysis (not reported here) incorporating reviews of Q3 and Q4 quartiles in the popular group and Q1 and Q2 in the non-popular group, in general, do not yield any distinguishing patterns. Therefore, in this study, we exclude reviews from the Q2 and Q3 quartiles to investigate whether a large margin between popular and non-popular categories indeed yields better distinguishing signals. The final dataset contains 2259 popular (all reviews from Q4 ) and 2555 non-popular reviews (all reviews from Q1). A few examples are shown in Fig \ref{examples}.

\section{Feature Analysis}

We consider various types of features in our analysis, including lexical, affective, discourse, and readability metrics that have been explored in prior studies for related research \cite{choi2020empirical,ghose2010estimating,diaz2018modeling}.

\subsection{Length Feature}
The length-related statistics provide insight into various textual aspects, such as the level of detail, the level of engagement, and the complexity of a review \cite{huang2015study}. We compute the distribution of the following length-related features: i) review length (\# sentences), ii) review length (\# words), and iii) sentence length (\# words) in popular and non-popular groups to determine the similarities and differences.

\subsection{Affective Feature}
Psychological features, specifically affective features in a text, encompass attributes related to emotions and sentiments. We examine the distributions of two types of affective features—sentiment and emotion—in two groups, as earlier studies have indicated associations with product helpfulness \cite{malik2017helpfulness}.

\paragraph{ \textbf{Presence of Sentiment}}
We scrutinize the sentiment in the reviews of two groups, such as the presence of sentiment and opinion words based on two popular English sentiment lexicons: Opinion Lexicon \cite{hu2004mining} and VADER \cite{hutto2014vader}. This assessment unveils how sentiments and opinions influence the popularity of user-provided feedback.

\paragraph{\textbf{Presence of Emotion}}
To recognize the presence of emotion in the text, we utilize the NRC Emotion Lexicon \cite{mohammad2010emotions}. The following six types of emotions (i.e., \textit{anger, anticipation, surprise, sadness, joy,} and \textit{disgust}) are investigated in the reviews of two groups.


\subsection{Lexical Feature}

\paragraph{\textbf{Presence of Capitalization}}
Capitalization is a common way to strengthen an expression or incorporate a louder tone within the text  \cite{pak2018value} and labels certain parts of an utterance as more salient than others \cite{chan2018social}.

\paragraph{\textbf{Presence of Common Words}}
The coverage of two common wordlists, one with 10,000 most common English words (large list) and the other with 5,000 most common English words (small list) are computed for the popular and non-popular review groups. \footnote {\url {https://github.com/sazzadcsedu/ReviewPopularity}}. The purpose is to find whether the presence of rarely used words inflicts any negative consequence on the popularity of the reviews.


\paragraph{\textbf{Lexical Diversity}}
Lexical diversity investigates the presence of unique words in a text corpus \cite{duran2004developmental}. We compute the ratios of distinct words in each review for both groups and then report the mean, median, and standard deviations. The goal is to find whether lexical diversity has any positive or negative influence on the popularity of the reviews. 





\subsection{Discourse Feature}

We further examine various discourse features, such as connectives and conjunctives, that are related to the coherence and cohesion of the text. 

\paragraph{\textbf{Connectives}}
The use of connecting words creates cohesive links between ideas and clauses and provides clues about text organization \cite{graesser2004coh}. We consider the following five types of connectives-

\begin{itemize}
     \item Adding (ADD): \textit{and, also, as well as, moreover, furthermore, besides, in addition}.

     \item Cause and Effect (CE): \textit{because, so, therefore, thus, consequently, as a result of.}



     \item Emphasizing (EMP): \textit{above all, in particular, especially, significantly, indeed, notably, most of all}.

     \item Illustrating (ILL): \textit{for example, such as, for instance, as revealed by, in the case of, as shown by}.


 \item Sequence (SEQ): \textit{next, then, first, second, finally, meanwhile, after}.
    
\end{itemize}


\paragraph{\textbf{Subordinating Conjunctions}}
In addition, we investigate the presence of subordinating conjunctions that are related to the complex sentence. Complex sentences are usually more difficult to process than simple sentences; nevertheless, they convey a clear and informative message. A list of 50 commonly used subordinating conjunctions is considered in this study \footnote{\url{https://github.com/sazzadcsedu/ReviewPopularity}}.





\subsection{Semantic Feature}

\paragraph{\textbf{Presence of Named Entity}}

A named entity (NE) represents a real-world object, such as the name of a person or organization. The three most common types of NE are person, location, and organization.
We analyze the presence of these three types of NE in the reviews, mentions of the character names (PER), geographical entities (GE) such as the mention of countries, cities, or similar things, and organizations (ORG). The spaCy \cite{spacy2} library is used to identify the presence of NE in the text.



\section{Readability}
Readability refers to the reading easiness and understandability of a piece of text. Readability is related to the cognitive effort that is needed by an individual to understand and comprehend a piece of text \cite{zakaluk1988readability}. A text with high readability improves the likelihood that the reader will clearly understand the thoughts and ideas conveyed. Previous studies yielded mixed findings regarding the relationship between readability and the effectiveness of product reviews. For instance, Liu et al. \cite{liu2015makes} found positive effects of readability on the usefulness of product reviews, whereas Hong et al. \cite{hong2017understanding} reported no significant association between these two factors. Thus, in this study, we aim to investigate whether readability influences the popularity of reviews in the creative domain. Two well-known readability tests, FRE and FKG, are applied to assess the reading easiness of the reviews and their connection with popularity. FRE provides readability scores on a scale from 1 to 100, where a higher score indicates that a chunk of English text is comparatively easier to read. The FKG level, on the other hand, is equivalent to the US grade level of education. 

\section{ Feature Significance Test and Classification}

\subsection{Mann-Whitney U Test for Significance}
In addition to providing quantitative statistics of various attributes, we perform the significance test to see whether the differences are meaningful for the likely candidates. We apply the Mann-Whitney U Test to determine if there is a significant difference between the two groups, popular and non-popular, based on individual attributes. The Mann-Whitney U Test is a nonparametric test that makes no assumptions about the distribution of data (e.g., normal distribution).  If the probability value (p-value) is less than the significance level (we use 0.05), then we reject the null hypothesis that there is no difference between the two groups and conclude that there is a significant difference between the two groups.


\subsection{N-gram Classification}
In addition to analyzing various linguistic features individually, we further investigate whether word n-grams show distinguishing capabilities when used as inputs of the classifiers. Four popular classical machine learning (CML) classifiers:  Logistic Regression (LR),  Support Vector Machine (SVM), K-NN (K-nearest Neighbor), and Gradient Boosting Tree (XGBoost), are employed. The default parameter settings of the scikit-learn library for various classifiers are used. For the K-NN classifier, the value of K is set to 5. In addition, we fine-tune two variants of transformer-based pre-trained language models: BERT \cite{devlin2018bert} and RoBERTa \cite{liu2019roberta}. A mini-batch size of 16 and a learning rate of 0.00002 are used. During the training, 20\% samples are utilized as a validation set. The Adam optimizer \cite{kingma2014adam} and cross-entropy loss are used as optimizer and loss functions, respectively. The training process runs for 3 epochs, and an early stopping criterion is employed. The performances of various classifiers are evaluated based on 5-fold cross-validation and utilizing precision, recall, macro F1 score, and accuracy metrics.


\section{Results and Discussion}

\begin{table}
\caption{Various length related features in the reviews of two groups}
\centering
\begin{tabular}{ccc}

\textbf{Attribute} & \textbf{Popular} & \textbf{Non-popular}  \\
\cline{2-3}

\textbf{} & median/mean/std. & median/mean/std. \\
\hline

\hline
\#Sentences / review & 18/24.60/23.08  & 11/16.13/18.21\\

\#Words / review & 332/441.64/398.18 & 194/290.59/316.83\\

\#Words / sentence & 17.82/17.95/7.45 & 16.96/18.02/8.22
\end{tabular}
\label{statistical-features}
\vspace{-4mm}
\end{table}

Table \ref{statistical-features} shows the distributions of various length-related features in two groups. We notice some differences in the review length of popular and non-popular groups. The mean and median word-level review length for the popular group is much higher (almost 1.6 times) than for the non-popular groups, with a median value of 332 versus 194. This observation suggests that longer reviews are likely to be more popular. However, we find that the sentence length is similar in both groups, which indicates the shorter or longer sentence has no noticeable impact on the popularity of the review.

\begin{table}[!ht]
\caption{Percentage of sentiment words in the reviews of popular and non-popular groups}
\centering
\begin{tabular}{cccc}

 \textbf{Sentiment} & \textbf{Sentiment} & \textbf{Popular} & \textbf{Non-Popular}    \\
\textbf{Lexicon} & \textbf{Orientation} & Median/Mean/Std. & Median/Mean/Std.
\\

\hline
Opinion & Negative & 3.30/3.45/2.06  & 3.0/3.29/2.46  \\

Lexicon & Positive  & 4/4.30/2.40  &4.2/4.91/3.55 \\
\hline
VADER  & Negative  &  2.8/ 2.97/1.95 &2.6/2.77/2.21  \\
  & Positive  & 5.0/5.37/2.60 & 5.2/5.83/3.55\\

\end{tabular}
\label{sentiment-statitics}
\end{table}

\begin{table}
\caption{Percentage of emotion words in the reviews of popular and non-popular groups}
\centering
\begin{tabular}{ccccccc}

\textbf{Emotion} & \textbf{Popular} & \textbf{Non-popular} \\

 & Median/Mean/Std. & Median/Mean/Std.
\\
\hline
\textbf{Anger} & 1.185/1.341/1.257 & 1.03/1.24/1.42 \\
\hline
\textbf{Anticipaton} & 2.286/2.473/1.610 &   2.22/2.54/2.12 \\
\hline
\textbf{Disgust} & 0.869/1.041/1.13 & 0.75/1.0/1.37 \\
\hline
\textbf{Joy}  &  1.8210/2.0610/1.6204 &  1.82/2.18/2.21  \\
 \hline
\textbf{Sadness} & 1.333/1.465/1.151 & 1.15/1.35/1.43 \\
\hline
\textbf{Surprise} &  0.984/1.126/1.088 &  0.94/1.17/ 1.38

\end{tabular}
\label{emotion-statitics}
\end{table}

\begin{table}
\caption{The presence of two sets of commonly-used English words (10,000 and 5,000) in the reviews of two groups}
\centering
\begin{tabular}{ccccc}
\textbf{\#Common Words} & \textbf{Group} & \textbf{Median}  & \textbf{Mean}  & \textbf{Std.}  \\
\hline
\hline
 10000 & Popular &   86.76\% & 86.11\% & 5.688\%\\
&Non-popular & 87.86\% & 87.34\% & 5.7\% \\
\hline
5000 & Popular &  73.03\% & 72.49\% & 6.60\%\\
&Non-popular & 74.00\% & 73.69\% & 6.80\% \\
\end{tabular}
\label{common_words}
\end{table}

\begin{table}
\caption{The percentage of unique and capitalized words in the reviews of two groups}
\centering
\begin{tabular}{ccccc} 

\hline
& \multicolumn{3}{c}{\textbf{Unique words}}  \\
\hline
\textbf{Group} & \textbf{Median} & \textbf{Mean} & \textbf{Std.} \ \\
\hline

Popular & 62.80\% & 64.33\% & 11.86\%\\
Non-popular & 67.86\% & 69.52\% & 13.29\% \\

\hline
& \multicolumn{3}{c}{\textbf{Capitalized words}} \\
\hline
&  \underline { \textbf{ (\%) of reviews}} & \underline {\textbf{(\%) in words (mean)}}
\\
 Popular & 787/2259 (34.83\%)  &  1.63\%\\
Non-popular &  559/2555 (21.87\%)  &  0.67\% 

\end{tabular}
\label{unique_words}
\end{table}

As we can see from Table \ref{sentiment-statitics},  the presence of the sentiment words is similar to both review groups based on mean and median values. We notice that this observation is true for both types of semantic orientations (positive and negative) and both lexicons (Opinion Lexicon and VADER). Table \ref{emotion-statitics} shows that, overall, the percentage of the emotion word is a bit higher in popular groups. Especially, the emotions related to Anger, Disgust, and Sadness are at least 10\% higher in the popular group than the non-popular group. 



\begin{table}
\caption{Presence of various types of connectives in the reviews of popular and non-popular groups (shown as percentage)}
\centering
\begin{tabular}{ccc}

\textbf{Connective} & \textbf{Popular} & \textbf{Non-popular} \\
\cline{2-3}
\textbf{} & \textbf{Median/Mean/Std.} & \textbf{Median/Mean/Std.} \\
\hline
ADD & 3.17/3.25/1.59 & 3.14/3.25/1.82  \\

CE & 3.52/3.63/1.82 &  3.7/3.86/2.29 \\
 
EMP & 0.0/0.05/0.15 &  0.0/0.06/0.31 \\

ILL & 0.0/0.03/0.1 & 0.0/0.02/0.12\\

SEQ & 0.54/0.68 /0.73 &  0.51/0.78/1.21 
 
\end{tabular}
\label{connectives_statitics}
\end{table}


From table \ref{common_words}, we find the percentages of common words (i.e., 5000 and 10000-word lists) are similar in both review groups. The results indicate that the presence of unfamiliar or less-known words does not show any relation to the likeability of the reviews. We notice that the presence of unique words is a bit higher in the popular group, which suggests that people tend to prefer reviews with fewer repetitive words. Similarly, the percentage of capitalized words (Table \ref{unique_words}) is higher in popular reviews. This observation suggests that strong feelings are somehow related to  the likability of reviews. 

We observe the percentages of connectives are similar to both popular and non-popular groups (Table \ref{connectives_statitics}). Among all the five types of connectives considered, none of them show any differences between the two groups. Similar patterns are observed for subordinate conjunctions (Table \ref{subordinate_conjunctions}), which suggest discourse-related features hardly provide any distinguishing signals to separate popular and non-popular groups.

\begin{table}
\caption{The presence of subordinating conjunctions in the reviews of popular and non-popular groups}
\centering
\begin{tabular}{cccc} 
\textbf{Review Group} & \textbf{Median}  & \textbf{Mean}  & \textbf{Std.}  \\
\hline
 Popular & 4.22\% & 4.15\% &  2.09\%  \\
Non-Popular &  4.17\% & 4.15\% & 1.76\%
\end{tabular}
\label{subordinate_conjunctions}
\end{table}

Regarding the presence of the named entity (Table \ref{named-entity}), although not significant, we observe some differences in their presence in popular and non-popular reviews. We find that popular reviews usually have more occurrences of all three types of entities. The readability tests, which indicate how easy it is to read a particular piece of text, show minimal differences between the two groups of reviews (Table \ref{readibility}), which implies that the readability measures are not a good indicator of the likeability of the review.

\begin{table}
\caption{The presence of named entities in the reviews of popular and non-popular groups}
\centering
\begin{tabular}{ccccc} 
\textbf{Entity}  & \textbf{Review Group} & \textbf{Median}  & \textbf{Mean}  & \textbf{Std.}  \\
\hline

 PER & Popular & 0.8\% & 1.02\% & 1.0\%\\
 & Non-Popular & 0.6\% & 0.9\% & 1.0\% \\
\hline

ORG &   Popular &  0.6\%& 0.77\% & 1.14\%\\
  &Non-popular & 0.3\% & 0.68\% & 1.02\% \\
\hline
GE  & Popular &  0.1\% & 0.32\% & 0.55\%\\
& Non-popular & 0.0\% &  0.28\% & 0.68\% \\

\end{tabular}
\label{named-entity}
\end{table}

\begin{table}
\caption{Statistics of FRE and FKG scores in popular and non-popular groups}
\centering
\begin{tabular}{ccccc}
\textbf{Readability}   & \textbf{ Group} & \textbf{Median}  & \textbf{Mean}  & \textbf{Std.}  \\
\hline
FRE  &Popular & 59.29 &  58.70 & 13.18\\
& Non-popular & 61.32 & 60.32 &  13.52 \\
\hline
FKG &  Popular & 9.47 & 9.78 & 3.58\\
 & Non-popular &  9.04 & 9.29 & 3.50
\end{tabular}
\label{readibility}
\end{table}


Table \ref{classifier} provides the precision, recall, macro F1 scores, and accuracy of various traditional  ML classifiers and transformers-based BERT and RoBERTa models. We find that classical ML algorithms provide macro F1 scores in the range of 0.56-0.63. All classifiers yield poor performance, the best F1 score obtained at 0.65 by transformer-based language models.

\begin{table}
\caption{Performance of various classifiers for determining review popularity using word n-gram features}
\centering
\begin{tabular}{ccccc}
 \textbf{Classifier} & \textbf{Precision} & \textbf{Recall}  & \textbf{Macro F1} &  \textbf{Accuracy}
\\

\hline

LR &0.63  &0.63  &0.63 & 0.63\\
SVM &  0.63 & 0.63 & 0.63 & 0.63  \\
XGBoost &  0.60 & 0.60 & 0.60 & 0.60 \\
K-NN & 0.56 & 0.55 & 0.56 & 0.54  \\
RoBERTa &  0.65 & 0.64 & 0.65 &  0.66 \\
BERT & 0.65 & 0.64 & 0.64 &  0.66
\end{tabular}
\vspace{-1.5em}
\label{classifier}
\end{table}

The analysis regarding the connection between various textual features and review likeability suggests that most of the features are unrelated to review likeability. In addition, the poor performance of word n-gram-based features reveals the limitation of word-based content features.

\subsection{Statistical Significance}
We further investigate which features exhibit significant differences between the two groups of book reviews. We employ the Mann-Whitney U test with a probabilistic p-value of 0.05  to discover the significant attributes. Based on the significance test, we notice the review length yields significant differences in the two groups having a p-value less than 0.05. Regarding the sentiment lexicon, we find that negative sentiment shows significant differences in the two groups, while positive sentiment does not. Additionally, the number of unique words differs significantly in the two groups. Regarding the discourse and semantic features, we observe no significant differences.



\subsection{Practical Implications}
The study investigates how user-generated text in book reviews becomes popular. Understanding these factors provides valuable insights into user behavior and preferences, benefiting authors, publishers, and online platforms in content creation. By incorporating these insights into marketing strategies, authors and publishers can effectively promote their books. Additionally, the study contributes to existing research with a comprehensive analysis of popularity factors, making it valuable for scholars interested in user-generated content, text analytics, and online communities. Overall, the study enhances our understanding of user behavior and content creation in the digital age.

\section{Summary and Future Work}
In this study, we explore various linguistic aspects of book reviews to understand how they are related to the attained popularity. We utilize a dataset of around 9000 book reviews to extract varied types of features and investigate their influences on the popularity of the reviews. We find, except for a few attributes like review length and the presence of capitalized and unique words, most of the other linguistic features exhibit similar patterns in popular and non-popular reviews. Besides, we observe that ML classifiers demonstrate inadequate efficacy for automatic review popularity determination, which reveals the challenges associated with this task. The results suggest that determining the popularity of reviews in the creative domain is highly challenging.

Despite predominantly negative results and findings, this study lays the groundwork for future analyses by highlighting the challenges and limitations of conventional attributes in distinguishing the likeness of user-generated content in creative domains. The future work will incorporate additional lexical, semantic, and grammatical features and an enlarged dataset for identifying better signals for review popularity projection.





\bibliography{anthology}
\bibliographystyle{ieeetr}
\end{document}